\documentclass[letterpaper, 10 pt, conference]{ieeeconf}  

\IEEEoverridecommandlockouts                              

\usepackage{graphicx} 
\usepackage{times} 
\usepackage{amsmath} 
\usepackage{amssymb}  
\usepackage{cite}
\usepackage{array,booktabs,makecell}
\usepackage{threeparttable}
\usepackage{tabularx}
\usepackage{svg}
\usepackage{adjustbox}
\usepackage{cellspace}
\usepackage{changes}
\usepackage[caption=false, font=footnotesize, farskip=0pt]{subfig}
\usepackage{hyperref}

\DeclareMathOperator*{\argmin}{arg\,min}

\title{\LARGE \bf
Challenges of SLAM in extremely unstructured environments: the DLR Planetary Stereo, Solid-State LiDAR, Inertial Dataset}

\author{Riccardo Giubilato$^{1}$, Wolfgang St\"urzl$^{1}$, Armin Wedler$^{1}$ and Rudolph Triebel$^{1,2}$
\thanks{$^{1}$German Aerospace Center (DLR), Institute of Robotics and Mechatronics, 82234 We\ss ling, Germany. 
        {\tt\small firstname.lastname@dlr.de}}%
\thanks{$^{2}$Department of Aerospace and Geodesy, Technical University of Munich, 80333 M\"unchen, Germany}
}

\DeclareUnicodeCharacter{2212}{\textendash}
\newcommand{\etnaone}{{\fontfamily{lmss}\selectfont s3li\_traverse\_1}}
\newcommand{\etnastraight}{{\fontfamily{lmss}\selectfont s3li\_traverse\_2}}
\newcommand{\etnacist}{{\fontfamily{lmss}\selectfont s3li\_crater}}
\newcommand{\etnanine}{{\fontfamily{lmss}\selectfont s3li\_loops}}
\newcommand{\etnacistinout}{{\fontfamily{lmss}\selectfont s3li\_crater\_inout}}
\newcommand{\etnamapping}{{\fontfamily{lmss}\selectfont s3li\_mapping}}
\newcommand{\etnalandmarks}{{\fontfamily{lmss}\selectfont s3li\_landmarks}}

\newcommand{\param}[1]{{\fontfamily{lmss}\selectfont #1}}

\newcommand{\revision}[1]{{\color{black}#1}}

\let\labelindent\relax
\usepackage{enumitem}
\newlist{tabitemize}{itemize}{1}
\setlist[tabitemize]{label=\textbullet,leftmargin=*,topsep=0ex,parsep=0pt,
                  after=\vspace{-\baselineskip},before=\vspace{-0.75\baselineskip}}

\begin{document}
\bstctlcite{IEEEexample:BSTcontrol}

\maketitle
\thispagestyle{empty}
\pagestyle{empty}

\begin{abstract}
We present the DLR Planetary Stereo, Solid-State LiDAR, Inertial (S3LI) dataset, recorded on Mt. Etna, Sicily, \revision{an environment analogous to the Moon and Mars}, using a hand-held sensor suite with attributes suitable for implementation on a space-like mobile rover. 
The environment is characterized by challenging conditions regarding both the visual and structural appearance: 
severe visual aliasing poses significant limitations to the ability of visual SLAM systems to perform place recognition, while the absence of outstanding structural details, joined with the limited Field-of-View of the utilized Solid-State LiDAR sensor, challenges traditional LiDAR SLAM for the task of pose estimation using point clouds alone.
With this data, that covers more than 4 kilometers of travel on soft volcanic slopes, we aim to: 1) provide a tool to expose limitations of state-of-the-art SLAM systems with respect to environments, which are not present in widely available datasets and 2) motivate the development of novel localization and mapping approaches, that rely efficiently on the complementary capabilities of the two sensors. 
The dataset is accessible at the following url: \url{https://rmc.dlr.de/s3li_dataset}

\end{abstract}

\section{INTRODUCTION}
Simultaneous Localization and Mapping (SLAM) techniques 
can nowadays be considered mature enough to find applications in many fields such as autonomous driving \cite{bresson2017simultaneous,burki2019vizard}, automated construction \cite{mascaro2021towards} or agriculture \cite{shu2021slam,oliveira2021advances}. 
While typically implemented with the use of multi-camera setups or RGB-D sensors, the recent democratization of 3D LiDAR sensors allowed easy integration of range sensing on mobile robots, enabling robust and large-scale mapping especially in urban or man-made scenarios. 

The wide variety of published visual or LiDAR SLAM approaches \cite{sharafutdinov2021comparison,dellenbach2021s,shan2018lego} might suggest that the SLAM problem could be considered solved. However, we argue that this is far from the truth, specifically when dealing with environments that are unstructured and severely aliased \cite{giubilato2020GPGM,gentil2020gaussian}. This is especially true in the case of planetary analogous environments, which are usually located in extreme locations such as deserts or volcanic surfaces. 
In this case, the lack of uniquely identifiable visual or structural features prevents from performing place recognition in a robust and repeatable manner.
Moreover, when dealing with LiDAR SLAM, the lack of vertical structures leads to missing constraints for a robust convergence of ICP-style algorithms \cite{gelfand2003geometrically}.  

Typical datasets fail to address this extreme case. Among the most popular, KITTI \cite{geiger2013vision}, Oxford RobotCar \cite{maddern20171}, KAIST \cite{choi2018kaist} and 4Season \cite{wenzel20204seasons} tackle the case of autonomous driving in urban scenarios. 
While presenting sequences with significant challenges, e.g., dynamic objects or large time intervals spanning seasons, 
the point of view of a road vehicle, as well as the panoramic 3D LiDAR scans captured within highly structured scenarios, facilitate the tasks of computing odometry and performing place recognition.

\begin{figure}[t]
    \centering
    {
      \footnotesize
      \includesvg[width=\linewidth]{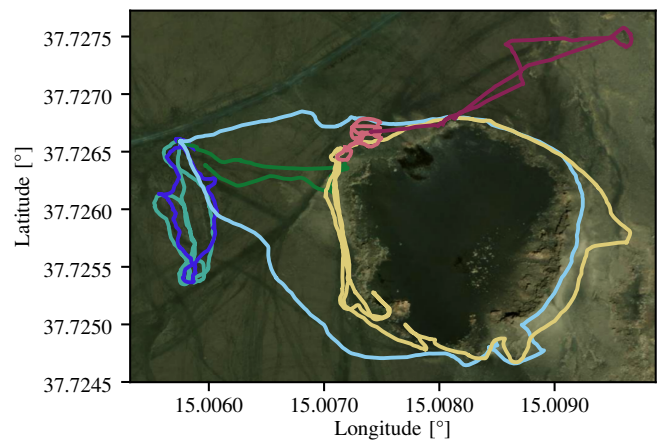}
    } \\ \vspace{.5cm}
    \vspace{-0.6cm}
    \caption{\footnotesize Bird's-eye view of the site where the dataset is recorded with trajectories for each sequence overlaid on top. The Cisternazza crater is visible on the right. 
    {
    Tiles (C) Esri -- Source: Esri, i-cubed, USDA, USGS, AEX, GeoEye, Getmapping, Aerogrid, IGN, IGP, UPR-EGP and the GIS User Community}}
    \label{fig:overview_all}
\end{figure}

Other datasets, such as TUM RGB-D \cite{sturm2012benchmark}, TUM VI \cite{schubert2018tum} or the ETH3D SLAM benchmark \cite{Schops_2019_CVPR} provide sequences recorded with hand-held stereo or RGB-D cameras principally in indoor environments. Their sequences tend to be short and frequently re-observing places from similar viewpoints and under very limited changes in the visual appearance.

To overcome the limitations related to the practical realization of the sensor setup, but also to model several motion characteristics and environments, many synthetic datasets have been recently released such as ICL-NUIM \cite{handa2014benchmark} or TartanAir \cite{wang2020tartanair}. While this has many advantages for the development of SLAM algorithms, e.g., the diversity of motion patterns and scenes is beneficial towards training of learning-based approaches, we believe that, especially for robots operating in the field, unstructured natural environments still offer the most challenging conditions for the implementation of visual or LiDAR-based SLAM. 

Several datasets have been proposed which are recorded from the perspective of a mobile robot operating in the field \cite{furgale2012devon}. The Chilean Underground Mine dataset \cite{leung2017chilean} provides sequences recorded in a mine with a stereo camera, a 3D LiDAR and a radar. The Visual-Inertial Canoe dataset \cite{miller2018visual} provides sequences where the main challenge is represented by reflections on the water. The Katwijk beach dataset \cite{hewitt2018katwijk} is recorded on a beach with artificial landmarks from a planetary-like rover equipped with a stereo camera that has similar characteristics to the one of the ExoMars rover. However challenging, the environment of caves and mines is, as was also shown during the DARPA SubT Challenge~\cite{rogers2020test,bouman2020autonomous,rouvcek2019darpa}, particularly suited for LiDAR-based SLAM, as are the stone replicas on the flat sandy surface of the Katwijk beach. 
The MADMAX dataset \cite{meyer2021madmax}, collected in the Morocco desert using a hand-held stereo unit with accurate ground truth, offers stereo sequences which are particularly challenging for traditional place recognition, but does not provide LiDAR data, which could be beneficial as it can provide information for localization invariant to the visual appearance. \revision{Similarly, \cite{vayugundla2018datasets} provides two sequences recorded on Mt. Etna with a mobile rover, which was only equipped with a stereo camera mounted on a pan-tilt unit.}

\begin{table}[]
    \centering
    \caption{SLAM datasets in unstructured environments}
    \begin{threeparttable}
        \renewcommand{\arraystretch}{1}
        \setlength\tabcolsep{5 pt}
        \begin{tabular}{c|ccccc}
            \textbf{Ref} & \textbf{Sensors} & \textbf{N {seq}} & \textbf{Length} & \textbf{GT} & \textbf{Obs}\tnote{4} \\ \hline\hline
            \cite{hewitt2018katwijk} & \makecell[c]{Stereo \\ LiDAR (3D)} & 3 & \makecell[c]{$\sim$ 2 km \\ $\sim$ 100 min} & D-GNSS & R \\ \hline
            \cite{leung2017chilean} & \makecell[c]{Stereo \\ RADAR (2D)} & 43\tnote{1} & \makecell[c]{$\sim$ 2 km} & LiDAR\tnote{2}& R \\ \hline
            \cite{vayugundla2018datasets} & \makecell[c]{Stereo} & 2 & \makecell[c]{$\sim$ 1 km} & D-GNSS & R \\ \hline
            \cite{furgale2012devon} & \makecell[c]{Stereo} & 23\tnote{3} & \makecell[c]{$\sim$ 10 km} & D-GNSS & R \\ \hline
            \cite{meyer2021madmax} & \makecell[c]{Stereo \\ Omni} & 35 & $\sim$ 9.2 km & \makecell[c]{D-GNSS \\ Heading} & H \\ \hline
            \textbf{Ours} & \makecell[c]{Stereo \\ LiDAR (Solid St.)\!\!} & 7 & \makecell[c]{$\sim$ 4.3 km \\ $\sim$ 90 min} & D-GNSS & H \\
            \hline \hline
        \end{tabular}
        \begin{tablenotes}
            \item[1] Sequences with traverses, other 44 are static LiDAR scans 
                \item[2] High-res LiDAR scans captured statically at 30-40 meters distance
                \item[3] Splits of a long traverse without revisiting places
                \item[4] R: Rover, H: Handheld sensor assembly
            \end{tablenotes}
    \end{threeparttable}
    \label{tab:comparison_datasets}
\end{table}

We fill this gap and present here a dataset collected on the volcanic surface of Mt. Etna, Italy, a planetary analogous environment characterized by extreme visual aliasing and lack of distinct structural features. The dataset has been collected with a hand-held sensor assembly that includes a stereo monochrome camera setup, a Solid State LiDAR and an IMU. 
\revision{The setup replicates the typical height and point of view of the sensors mounted on a mobile robot. Although influenced by the walking pattern, the recordings look smooth and free of sharp or high-frequency motions.}
Seven sequences, 8 to 30 minutes long and covering distances up to 1.3 kilometers, have been recorded in order to evaluate the ability of SLAM algorithms: 
\begin{enumerate}
\item to accurately estimate odometry under harsh lighting conditions and lack of unique features,
\item to rely on different sensing modalities,
depending on the structure and appearance of the environment,
\item to use visual and structural information to robustly perform place recognition under severe aliasing.
\end{enumerate}

In summary, this paper presents the following contributions:
\begin{itemize}
    \item We release a dataset to challenge SLAM algorithms in an natural, extremely unstructured environment, that resembles the conditions of planetary-like scenarios.
    \item We evaluate a selection of state-of-the-art visual, and visual-inertial SLAM algorithms.
    \item We provide examples of using the data in unconventional manners, leveraging the interaction of cameras and LiDARs towards more robust SLAM in severely unstructured scenes.
\end{itemize}

\section{THE DATASET}
\begin{figure*}[t]
    \centering
    {   
        \renewcommand{\seriesdefault}{\bfdefault}
        \small
        \includesvg[width=\linewidth]{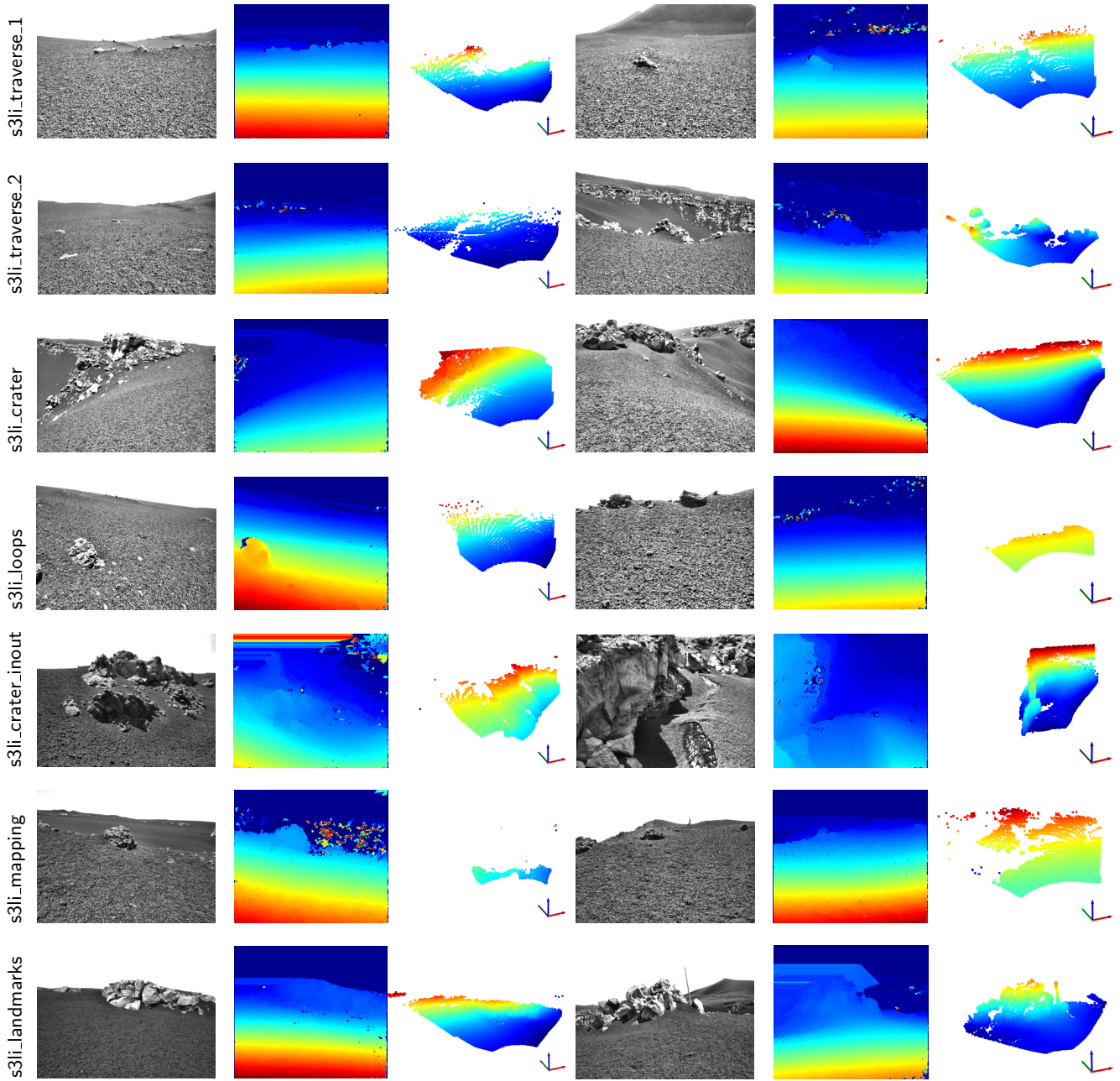}
    }
    \caption{Collection of shots from the 7 sequences of the dataset. For each sequence, two examples are shown that include an image from the left camera, the stereo disparity and the closest LiDAR scan in time. All images are shown after applying CLAHE (Contrast-Limited Adaptive Histogram Equalization), to enhance the contrast, and the disparity images are produced using OpenCV's SGBM with \param{numDisparities = 96} and \param{blockSize = 11}.}
    \label{fig:overview}
\end{figure*}
The dataset was recorded on the volcanic environment of Mt. Etna, Sicily, an active stratovolcano, which is denoted as a planetary analogous site for Moon and Mars due to its peculiar geological features \cite{preston2012concepts}. 
The dataset includes 7 sequences, that were recorded on a slope at 2650 meters above sea level and in proximity to the Cisternazza crater (see Fig.~\ref{fig:overview_all}).
Each sequence is located far from hiking trails and provides views of an extreme environment characterized by a surface of smooth dark lava ash, due to the frequent eruptions that happened overnight at the time of recording the data, extreme visual contrast, due to the darkness of the soil and the brightness of the sky, and scarcity of geological features that might facilitate localization with respect to prior observations of the same places. 
In comparison with other similar datasets, listed in Table~\ref{tab:comparison_datasets}, we provide an honest multi-sensory look at an environment, analogous to potential sites for robotic planetary exploration, without alterations, e.g. by introducing artificial landmarks, and without a prior selection based on properties that help specific sensing modalities, i.e., flatness of the soil, presence of outstanding structural details, or presence of distinct visual textures on the ground.  

\subsection{Sequences}

\begin{table*}[t]
    \centering
    \caption{Overview of the sequences}
    \begin{threeparttable}
        \begin{tabularx}{\linewidth}{l|c|c|X}
             \textbf{Name} & \textbf{Length} & \textbf{D-GPS Track\tnote{1}} & \textbf{Description} \\ 
             \hline\hline 
             \etnaone
             & 
                \begin{tabular}{l}\makecell[c]{371 m \\ 9 min}\end{tabular} 
             & 
             \adjustbox{valign=t}{
                \includesvg[height=3cm]{fig/GPS_tracks/bitmapped/etna_1}}
             &
             \begin{tabitemize}
                \item Traverse along a ridge of stones on a slight downhill. The environment is nicely contrasted, however the severe aliasing undermines the chances for visual place recognition. 
                \item The path intersects briefly in the middle section but there are no evident loop closure opportunities, except at the very end. 
                \item Challenging LiDAR clouds mainly due to the slope and the absence of distinct structures. 
             \end{tabitemize}
             \\ \hline
             \etnastraight & \makecell[c]{300 m \\ 8 min} & \adjustbox{valign=t}{
                \includesvg[height=2cm]{fig/GPS_tracks/bitmapped/etna_straight_mapping}}
             & 
             \begin{tabitemize}
                 \item Short traverse from the starting point to the rim of the Cisternazza crater and back.
                 \item Several panning motions during the traverse help to cover more area for purposes of mapping. There are basically no loop closure opportunities. 
                 \item As for \etnaone, LiDAR clouds have limited extent and do not include many useful structures.
             \end{tabitemize}
             \\ \hline
             \etnacist & \makecell[c]{1010 m \\ 19 min}
             & 
             \adjustbox{valign=t}{
                \includesvg[height=3cm]{fig/GPS_tracks/bitmapped/etna_cisternazza}}
             & 
             \begin{tabitemize}
                \item First long sequence around the rim of the crater, returning to the starting location. 
                \item The visual appearance is diverse, as many different terrain types are observed including sandy slopes and the rocky walls within the crater rim. 
                \item Opportunities for mapping the insides of the crater using the LiDAR, more predominant structures are present, although none can be used to close loops as they are not revisited.
             \end{tabitemize}
             \\ \hline
             \etnanine & 
             \makecell[c]{587 m \\ 13 min} 
             & 
             \adjustbox{valign=t}{
                \includesvg[height=3cm]{fig/GPS_tracks/bitmapped/etna_9}}
             & 
             \begin{tabitemize}
                \item Loopy sequence around the rocky ridges observed also in \etnaone\ and \etnastraight, plenty of opportunities for visual loop closure detection and/or for visual or structure-based segmentation of natural features.
                \item Less monotonous visual appearance than \etnaone\ and \etnastraight, a few salient rocky features are observed in "orbiting" motions.
                \item Structure is less interesting than visual appearance. 
             \end{tabitemize}
             \\ \hline
             \etnacistinout & \makecell[c]{1338 m \\ 28 min} 
             & 
             \adjustbox{valign=t}{
                \includesvg[height=3cm]{fig/GPS_tracks/bitmapped/etna_cisternazza_inout}}
             & 
             \begin{tabitemize}
                \item Long traverse around the crater, that also includes a short travel below the rim to observe some geological formations. They are revisited after completing the first traverse, offering opportunities for loop closure detection. 
                \item Challenging visual appearance characterized by sharp changes in illumination due to the exploration of the insides of the crater, often lying in the shadows
                \item Interesting sequence to map geological formations with the LiDAR, to be exploited also towards structure-based place recognition.
             \end{tabitemize}\\ \hline
             \etnamapping & \makecell[c]{242 m \\ 8 min} & \adjustbox{valign=t}{
                \includesvg[height=3cm]{fig/GPS_tracks/bitmapped/etna_mapping_test}}
             & 
             \begin{tabitemize}
                \item Sequence of limited spatial extent targeted at mapping, where the sensors observe the environment to cover as much area as possible. The terrain is characterized by ash slopes with few small boulders.
                \item Plenty of loop closure opportunities are available, however the contrast between the dark lava ash and harsh lighting challenges traditional visual methods.
                \item LiDAR measurements cover a dense area and can be interesting for complementing the mapping ability of the stereo camera as well as for providing solid ground constraints in the context of multi-modal SLAM.
             \end{tabitemize}
             \\ \hline
             \etnalandmarks & \makecell[c]{482 m \\ 9 min} & \adjustbox{valign=t}{
                \includesvg[height=2cm]{fig/GPS_tracks/bitmapped/etna_landmarks}}
             &
             \begin{tabitemize}
                \item Sequence of traverses between large and isolated rock formations, some of which are re-observed to allow for an easy recognition of structures and visual features.
                \item As for \etnamapping, the visual appearance is challenging due to the harsh contrast between dark lava ash and the bright sky. During the traverses, the ground offers limited visual details.
                \item The rock formations are observed from many viewpoints allowing for a complete reconstruction. Salient structures can easily be segmented for semantic mapping or matching.
             \end{tabitemize}
             \\ \hline\hline 
        \end{tabularx}
        \begin{tablenotes}
            \item[1] The green color denotes D-GNSS estimates under RTK-FIX ($\varepsilon\sim1\text{cm}$), while the yellow color denotes RTK-FLOAT ($\varepsilon\sim\lambda_\text{carrier}$).  
        \end{tablenotes}
    \end{threeparttable}
    \label{tab:sequences}
\end{table*}
The 7 sequences provided can be categorized according to the challenges they pose for localization algorithms. \etnaone, \etnastraight\  and \etnacist\ test the ability of localization methods or SLAM algorithms to provide accurate pose estimation during traverses. The first two observe mainly ash slopes, with the sporadic presence of rocks or boulders, while the third is a long traverse, that starts and ends at the initial location, around the rim of the crater, providing long range features for which depth is unknown. \etnanine\ and \etnacistinout\ are, instead, recorded with the purpose of evaluating the ability of SLAM algorithms to perform place recognition, exploiting either the visual or structural similarity. While the first revisits mainly unstructured slopes, that might be recognized through carefully tuned visual loop closure detectors, the second includes short repeating traverses within the rim of the crater, that could be exploited with both sensing modalities. \etnamapping\ instead tests the ability of SLAM algorithms to produce a complete and consistent map of the environment relying on stereo as well as LiDAR depth measurements, as a confined area is explored multiple times and there should be plenty of loop closure opportunities. Finally, \etnalandmarks\ allows testing of segmentation techniques for natural landmarks, with the aim of either semantic mapping or segment-based loop closure detection \cite{cramariuc2021semsegmap}. An overview of the sequences, including D-GNSS tracks, is provided in Table~\ref{tab:sequences}.

\subsection{The hand-held setup (S3LI)}
\begin{figure}[t]
    \centering
    \includegraphics[width=\linewidth]{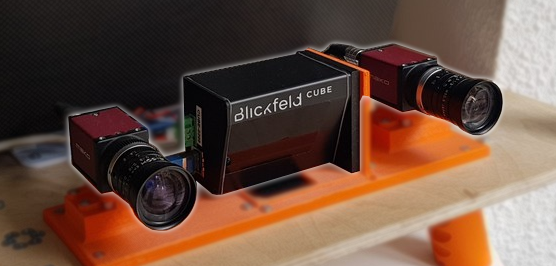}
    \caption{The Stereo, Solid-State LiDAR, Inertial (S3LI) sensor setup}
    \label{fig:setup}
\end{figure}
The sensor setup used to collect the data, depicted in Fig.~\ref{fig:setup}, is designed to be compliant to a potential implementation on a planetary-like rover \cite{schuster2019towards} and comprises: \\
\textbf{Stereo Camera}: two AVT Mako cameras are configured in a stereo setup with baseline 20 centimeters. The cameras are triggered at a frequency of 30 Hz to capture monochromatic images with resolution of $688 \times 512$ pixels using automatic exposure control to cope with the harsh lighting conditions on the mountain. 
\revision{Their timestamps are synchronized to the system time using the PTP protocol, where cameras function as slave to the master clock of the laptop.}
Compared to other datasets based on stereo images \cite{vayugundla2018datasets,meyer2021madmax}, we do not include depth images, as the computation of disparity is not intended as a sensor output, but as part of the utilization of the stereo images. Many SLAM architectures \cite{campos2021orb,usenko19nfr,Geneva2020ICRA}, in fact, although relying on a stereo image stream, do not make use of dense disparity images, but instead perform sparse feature matching when needed. 
Nevertheless, as an example and for the purpose of visualization, Fig.~\ref{fig:overview} presents a collection of disparity images obtained with OpenCV's SGBM stereo algorithm. 
Due to the automatic exposure control that, in combination with the extreme intensity differences between the dark soil and the bright sky, often compresses the ground details into dark patches, we recommend the usage of automatic contrast enhancement algorithms such as OpenCV's CLAHE. \\
\textbf{LiDAR}: we use a Blickfeld Cube-1 LiDAR, which employs a MEMS-actuated beam deflection mirror, instead of traditional 360$^\circ$ LiDARs, as this type of construction is more suited for potential space applications due to increased mechanical robustness and reduced weight and power consumption. For all sequences, the LiDAR is configured to capture point clouds with a maximum number of 17400 points with a Field-of-View of about 70$^\circ$H~$\times$~30$^\circ$V, which results in a scan rate of 4.7 Hz. Each data point is timestamped with respect to the origin of the scan and contains additional intensity information that can be used, for instance, to  identify low-quality measurements. \revision{The synchronization of the scan timestamps with respect to the time of the laptop is realized with NTP.}\\
\textbf{IMU}: an XSens MTi-G 10, \revision{connected via USB, }provides linear acceleration and angular velocity at a rate of 400 Hz. \\
\textbf{GPS receiver}: an Ublox f9p GNSS receiver mounted on the hand-held setup provides accurate differential estimates in conjunction with a base station. The data logs are processed in a later stage using the RTKLIB to obtain ground truth positions at a frequency of 5 Hz and centimeter-level accuracy. 
\begin{figure}[!t]
    \centering
    \subfloat{\includegraphics[width=.49\linewidth]{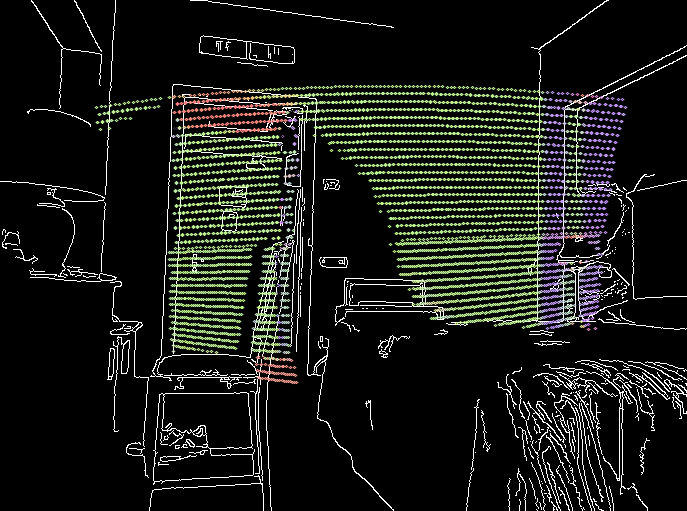}\label{<figure1>}} \hfill
    \subfloat{\includegraphics[width=.49\linewidth]{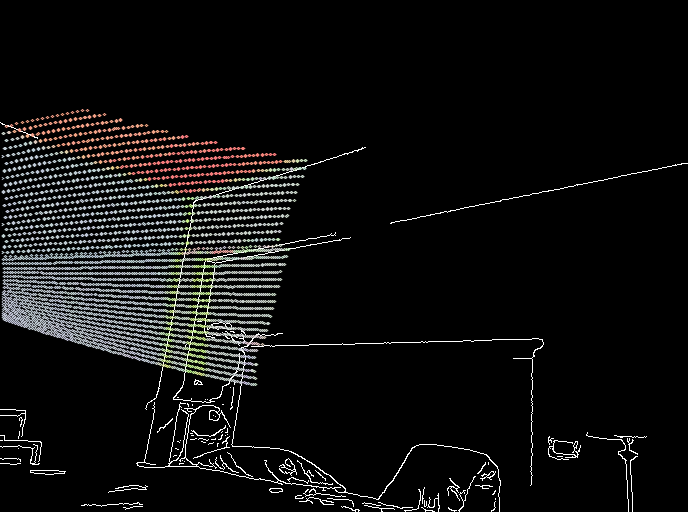}\label{<figure2>}} \\[1mm] 
    \subfloat{\includegraphics[width=.49\linewidth]{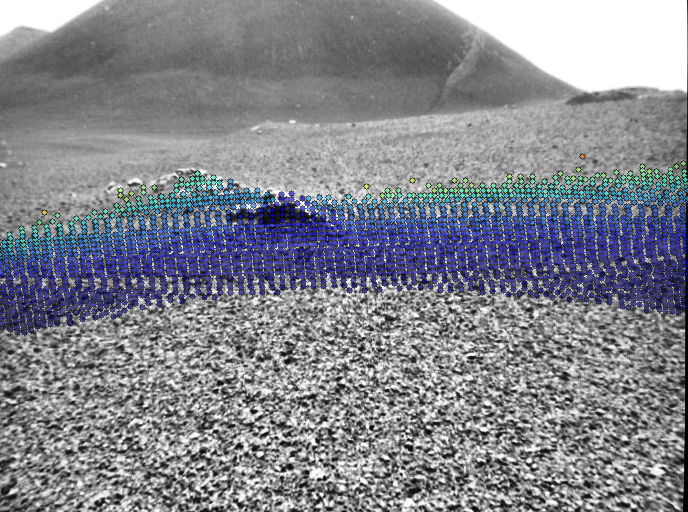}\label{fig:ex_lid_ov_e1_0}} \hfill
    \subfloat{\includegraphics[width=.49\linewidth]{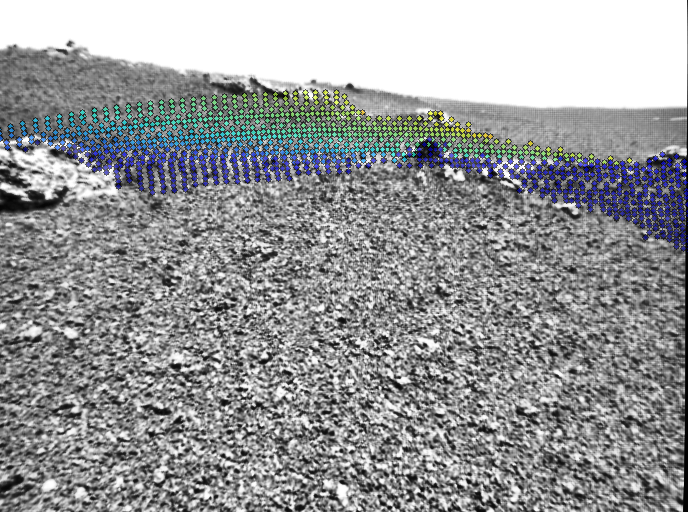}\label{fig:ex_lid_ov_e1_1}} \\[1mm] 
    \subfloat{\includegraphics[width=.49\linewidth, trim=0 0 0.1cm 0, clip]{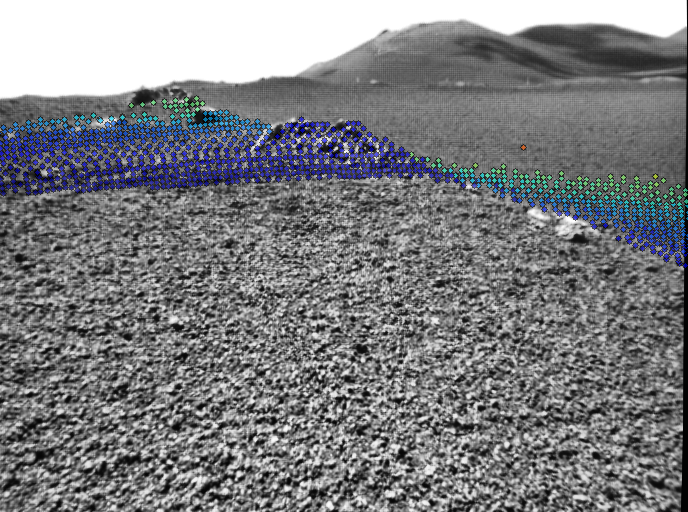}\label{<fig:ex_lid_ov_em_0}} \hfill
    \subfloat{\includegraphics[width=.49\linewidth, trim=0 0 0.1cm 0, clip]{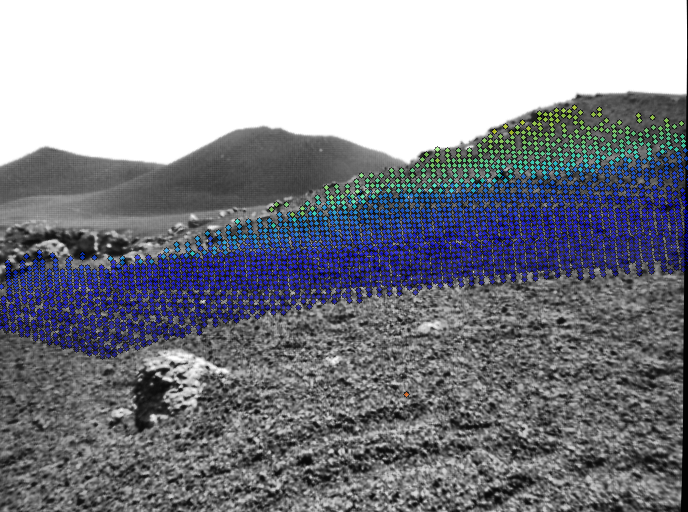}\label{fig:ex_lid_ov_em_1}}\\[1mm]
    \subfloat{\includegraphics[width=.49\linewidth]{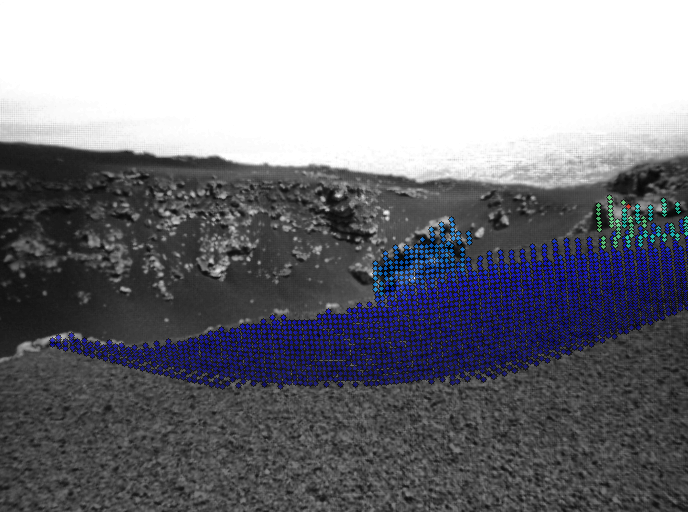}\label{<fig:ex_lid_ov_ecio_0}} \hfill
    \subfloat{\includegraphics[width=.49\linewidth]{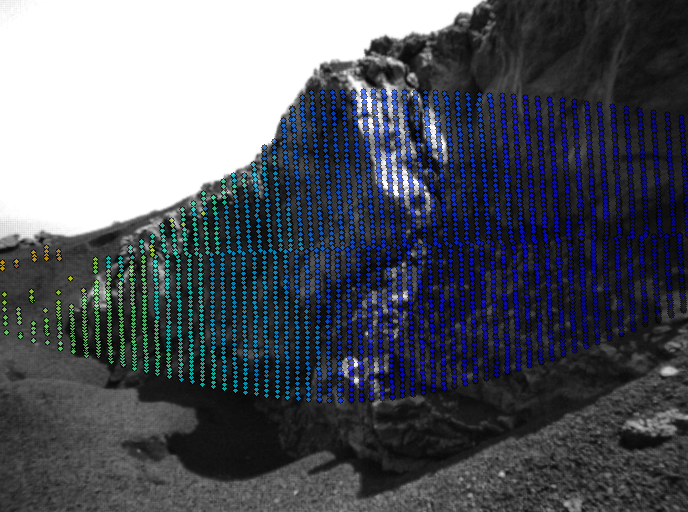}\label{fig:ex_lid_ov_ecio_1}}\\[1mm] 
    \subfloat{\includegraphics[width=.49\linewidth, trim=0 0 0.3cm 0, clip]{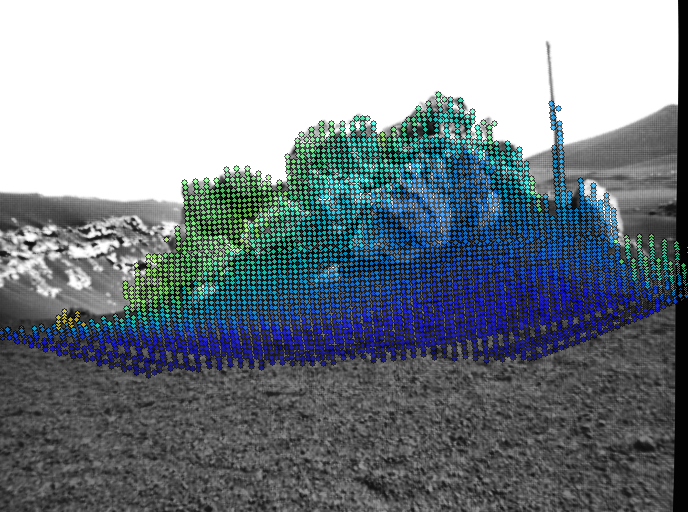}\label{<fig:ex_lid_ov_el_0}} \hfill
    \subfloat{\includegraphics[width=.49\linewidth, trim=0 0 0.3cm 0, clip]{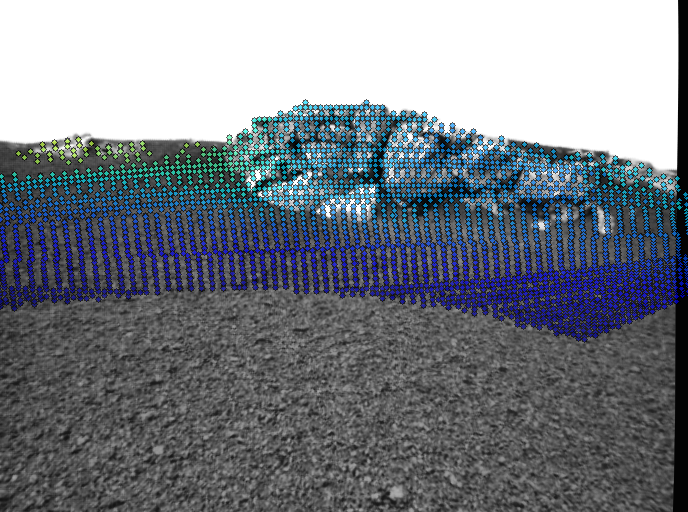}\label{fig:ex_lid_ov_el_1}}
    \caption{Alignment of LiDAR and (left) camera reference frames. In the top row: examples of binary edge images and projected LiDAR scans after extrinsic calibration. The color scheme of the projected LiDAR points reflects the direction of the normal vectors. The other rows show examples of LiDAR scans, projected on the left camera image, respectively from sequences \etnaone, \etnamapping\ and \etnacistinout. Point clouds are downsampled to avoid cluttering the images and color-coded according to the depth of points.}
    \label{fig:ex_lidar_calib}
\end{figure}
\subsection{Sensor calibration}
\begin{table*}[ht]
    \centering
    \begin{threeparttable}
        \caption{RMSE\tnote{*} (normalized) and ratio of completion of the compared SLAM algorithms}
        \setlength\tabcolsep{5 pt}
        \begin{tabular}{l|ccccccc}
              \textbf{Algorithm}&       \textbf{\etnaone} &   \textbf{\etnanine} & \textbf{\etnacist} & \textbf{\etnacistinout} &  \textbf{\etnalandmarks} &  \textbf{\etnamapping} & \textbf{\etnastraight} \\
        \hline\hline
              ORB-SLAM3 (S) \cite{campos2021orb} & 0.86 (100) &         \textbf{0.21} (100) &           0.87 (56.5) &          0.19 (69.5) &   - &           0.71 (41.2) &         0.38 (84.1) \\
              VINS\_Fusion (S) \cite{qin2019a} & 1.41 (100) &         6.23 (100) &          2.25 (100) &                 - &        3.25 (50.1) &          2.01 (62.5) &          1.33 (100)  \\
              VINS\_Fusion (SI) \cite{qin2019a}  &  0.62 (100) &                - &                  - &           \textbf{0.25} (100) &        \textbf{0.36} (100) &        - &     1.90 (100) \\
              OPEN\_VINS (SI) \cite{Geneva2020ICRA} & \textbf{0.54} (100) &         0.41 (80.3) &           0.54 (42.6) &          0.16 (86.9) &        0.77 (83.9) &     - &     0.39 (100)  \\
              BASALT (S) \cite{usenko19nfr} & 0.75 (100) &	0.35 (100) & 2.81 (100) &	1.08 (100) &	1.34 (100) & 0.47 (100) &	0.57 (100)	 \\
              BASALT (SI) \cite{usenko19nfr} & 0.66 (100) & 0.46 (100) & \textbf{1.64} (100) & 0.41 (100) & 0.86 (100) & \textbf{0.44} (100) & \textbf{0.38} (100) \\
        \hline \hline
        \end{tabular}
        \begin{tablenotes}
            \item[*] In bold are highlighted the best results, from those algorithms that completed the sequence
        \end{tablenotes}
    \end{threeparttable}
    \label{tab:SLAM_comparison}
\end{table*}

\textbf{Stereo camera and IMU}: The intrinsic and extrinsic parameters of the stereo camera pair are calibrated using Calde/Callab \cite{callab}, while the IMU to camera calibration is performed with Kalibr \cite{furgale2013unified}. In order to facilitate the playback of data, the stereo images are provided already rectified. In fact, as the lenses do not have a particularly wide field of view, the images corrected for distortion do not lose a significant amount of visual detail. \\
\textbf{LiDAR-Camera extrinsic calibration}: the LiDAR is extrinsically calibrated with respect to the left camera. As the solid-state LiDAR that we employed is characterized by quite a large minimum measurable depth of about 4 meters, traditional techniques for LiDAR to camera calibration proved either non-effective, e.g., in the case of a checkerboard pattern printed on planar target \cite{zhou2018automatic}, or too impractical to perform on the field, e.g., methodologies that rely on non-planar geometries  \cite{gong20133d,toth2020automatic}. For this reason, the transformation between the camera and LiDAR reference frames is determined by fixing the translation component to the one obtained by a CAD model of the sensor assembly, while the rotation component is optimized in an automatic edge alignment scheme \cite{levinson2013automatic,castorena2016autocalibration}. More specifically, $n$ pairs of left camera images and LiDAR scans are manually selected from recording of an indoor environment, characterized by structural regularity and scarcity of evident textures. From each image, strong Canny edges are selected to obtain binary images, where the identified edges are an input to the optimization scheme. From each corresponding LiDAR scan, we estimate the local curvatures, using the PCL library, and edge points, which lie in correspondence to depth discontinuities between neighboring points in the sequence of measurements of the LiDAR beam. Formally, the extrinsic calibration is solved as a non-linear least squares problem, where the goal is to find the optimal rotation between LiDAR and left camera, that aligns the projections of LiDAR points, lying on edges or discontinuities, to edges in the image:
\begin{equation}
    \argmin_{R^{\text{C}_\text{l}}_{\text{L}}} \sum_{i=1}^{n}\sum_{k=1}^{m_i} \omega_{ik} \rho_c\big( \| \mathbf{x}_{ik} - \pi\big(R^{\text{C}_\text{l}}_\text{L} (\mathbf{p}^\text{L}_{ik}-\mathbf{t}_{\text{C}_\text{l}}^{\text{L}})\big) \|\big),
\end{equation}
where $n$ is the number of image-scan pairs, $m_i$ is the number of correspondences for the $i$-th pair, $\pi$ is the camera projection function, $\mathbf{p}^\text{L}_{ik}$ and $\mathbf{x}_{ik}$ are, respectively, 3D LiDAR  and 2D image points on edges, $R^{\text{C}_\text{l}}_\text{L}$ and $\mathbf{t}_{\text{C}_\text{l}}^{\text{L}}$ are the rotation (to be optimized) and translation (known form CAD model) of the LiDAR in the left camera reference frame, $\rho_c$ is a Cauchy robust loss function, and $\omega_{ik}$ is a weight proportional to the local curvature of the point cloud. The association between LiDAR and image edge points is performed iteratively in a nearest-neighbor approach, and the problem is solved a few consecutive times reducing the value of parameter $c$ for the Cauchy loss. 

\section{USE-CASES}
\begin{figure*}[!t]
    \centering
    \subfloat[\etnaone]{\includegraphics[height=3.5cm]{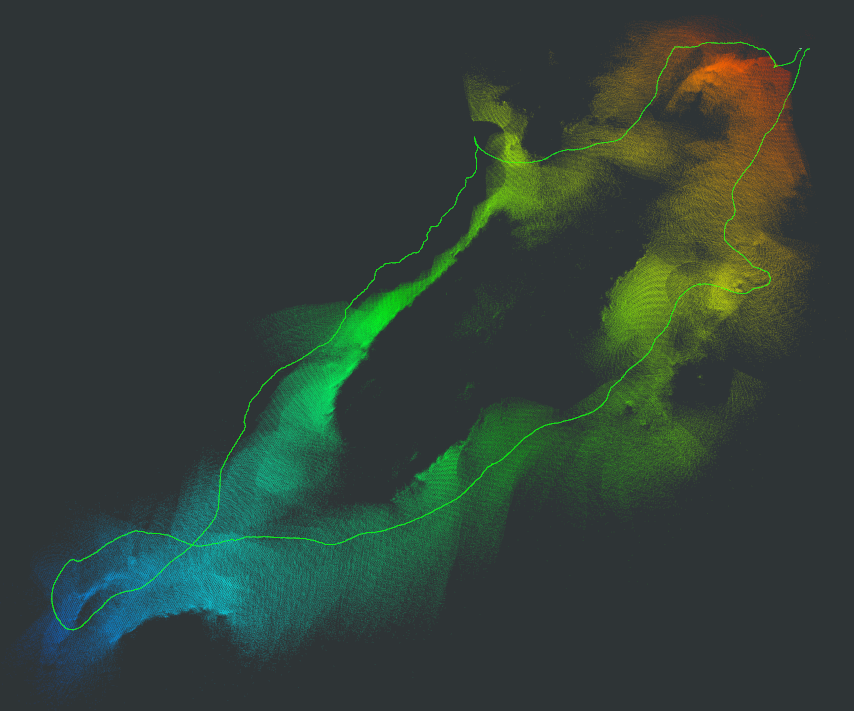}} \hfill
    \subfloat[\etnalandmarks]{\includegraphics[height=3.5cm]{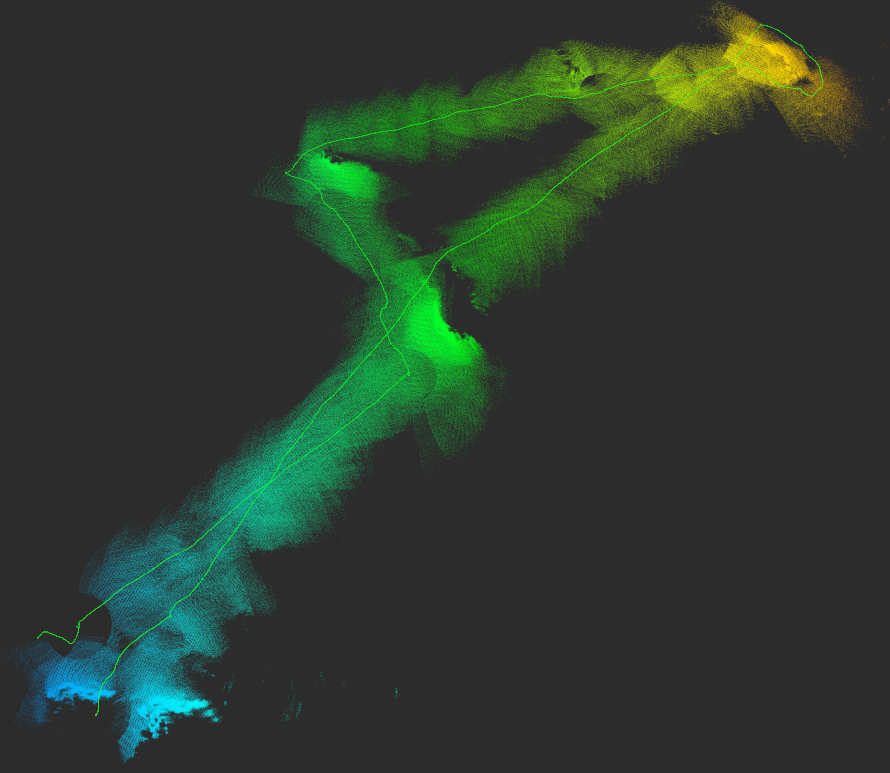}} \hfill
    \subfloat[\etnastraight]{\includegraphics[height=3.5cm]{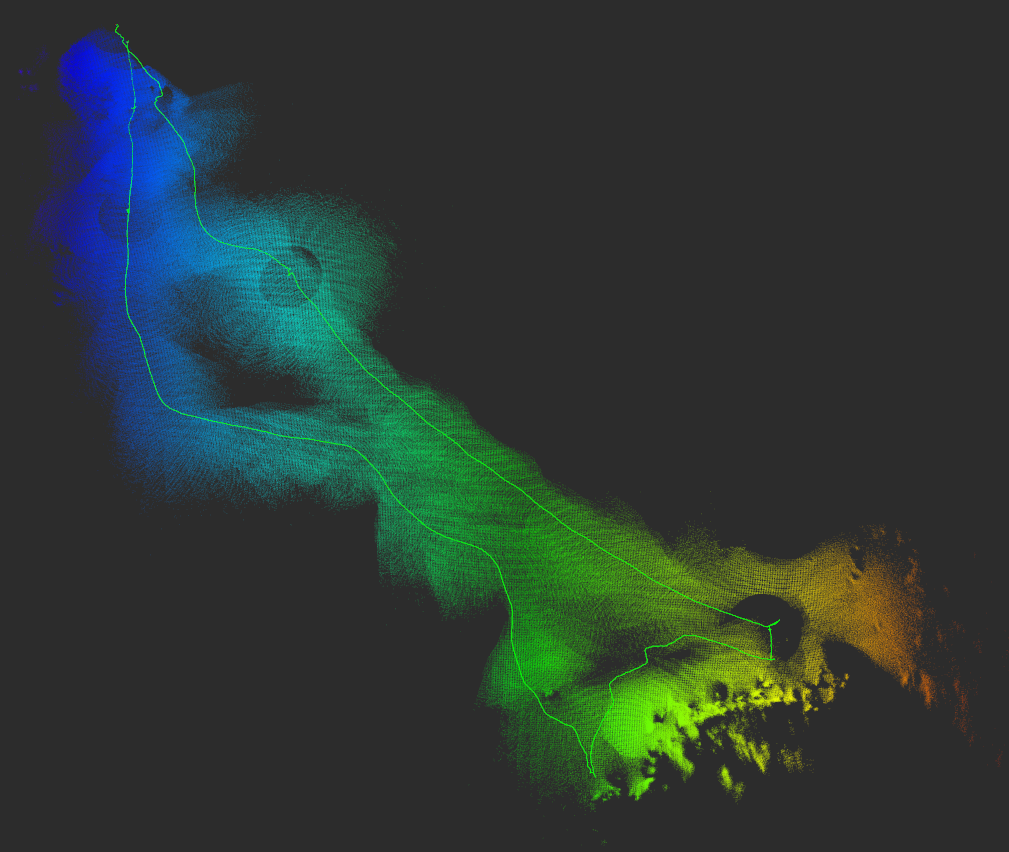}} \hfill 
    \subfloat[\etnacistinout\ (detail)]{\includegraphics[height=3.5cm]{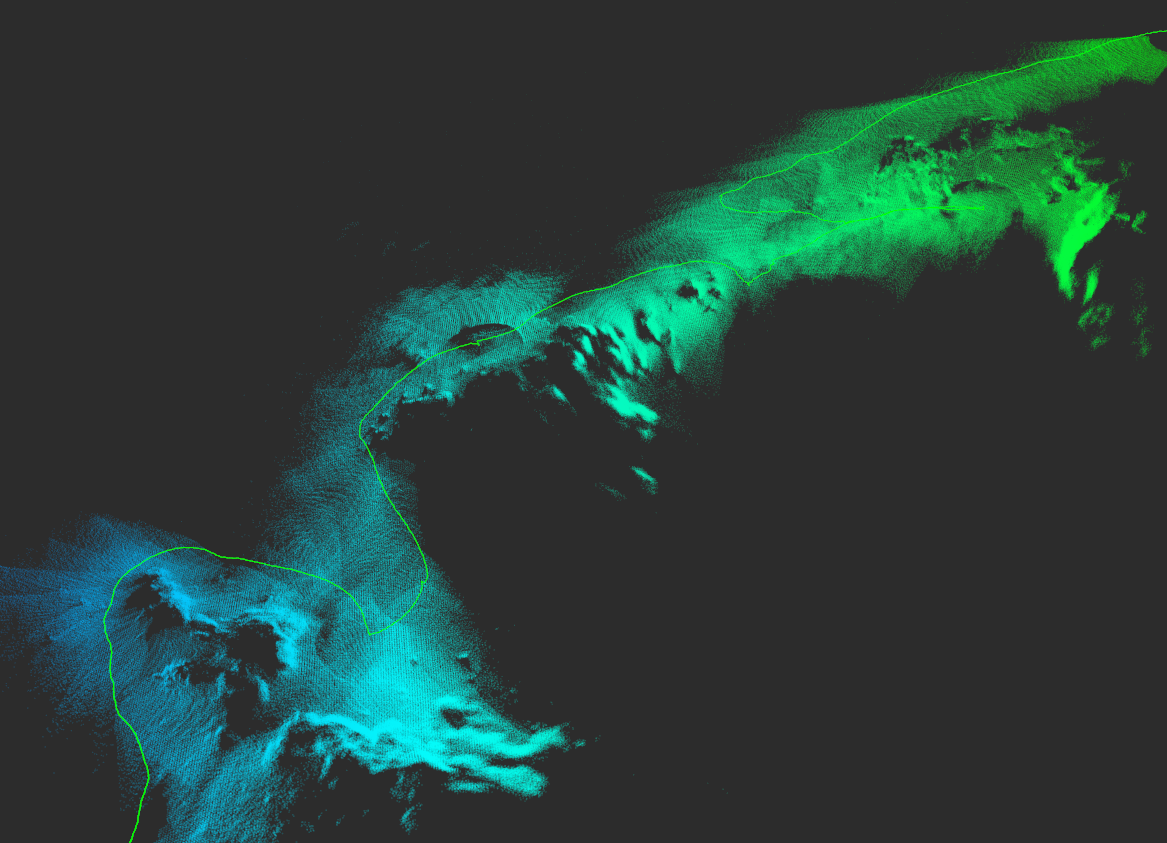}}
    \caption{Dense maps produced by assembling LiDAR point clouds using pose estimates of OPEN\_VINS, shown as a thin green line. (a-c) show maps produced on full sequences, the extent of the trajectories are visible in Table~\ref{tab:sequences}. (d) shows a detail from the map of \etnacistinout, that highlights the inner rim of the Cisternazza crater and includes several examples of rock formations.}
    \label{fig:example_lidar_mapping}
\end{figure*}
\subsection{Visual SLAM}
A major application area of this dataset is to evaluate the performances of SLAM algorithms in environments with challenging visual appearance. While the texture of the terrain can ease the task of, for instance, sparse optical flow, feature matching and place recognition are particularly challenged by the severe visual aliasing. 
The algorithms compared here are ORB\_SLAM3 \cite{campos2021orb}, VINS\_Fusion \cite{qin2019a}, OPEN\_VINS \cite{Geneva2020ICRA} and BASALT \cite{usenko19nfr} in stereo or stereo-inertial configuration. 
The estimated path is correlated to the D-GNSS ground truth using time correspondences. After this step, we align all pairs of matching positions using Horn's algorithm, and compute the root mean square of all the residuals. 
To be comparable amongst all sequences, the RMSE is normalized by the length of the associated ground truth trajectory.
Furthermore, as the RMSE is affected by the length of the estimated trajectory (shorter trajectories exhibit less translational and rotational drift) we provide, as an additional performance indicator, the proportion of each sequence for which poses are estimated from Visual SLAM. This value is computed as the length of the ground truth, associated to SLAM estimates, over its total length. Table~\ref{tab:SLAM_comparison} reports these metrics for all algorithms and for every sequence of the dataset. 
Compared to the others, ORB\_SLAM3, where camera tracking depends heavily on ORB descriptors matching, exhibits very good accuracy in all datasets, but frequently fails to estimate the whole sequence, especially when the observed environment is characterized almost completely by a very repetitive terrain, without the presence of rocks or features of unique appearance. The other algorithms, where the localization front-end depends on tracking strong corners through optical flow, do not suffer from this specific problem, but are generally more prone to pose drift.

As expected, the lowest errors are scored from 
stereo-inertial configurations, as the inertial measurements help significantly to deal with high accelerations and rotational velocities, typical of the walking motion patterns when recording with hand-held devices. However, compared to their stereo counterparts, the delicate procedures to initialize IMU parameters leads often to lose significant portions of the initial parts of the sequences. Furthermore, we experienced random behaviours of this step, especially with VINS\_Fusion and OPEN\_VINS, often initializing the state estimation at very different points after the beginning of sequences. For this reason, Table~\ref{tab:SLAM_comparison} reports the best results obtained during different runs of the same sequences, and omits results for which less than 30\% of the trajectory was estimated.

Overall, there is no apparent winner among all the algorithms compared, in terms of lowest RMSE values, although for robustness, indicated by the proportions of sequences successfully processed, the approach of BASALT would be the most successful.
Hence, this dataset can help the development of Visual SLAM approaches, considering the challenges provided by this type of environment, which are often missing in traditional datasets for Visual SLAM.

\subsection{LiDAR Mapping}
As Figs.~\ref{fig:overview} and \ref{fig:ex_lidar_calib} show, each LiDAR scan covers a small portion of the environment, which is located in front of the sensor setup within a relatively narrow Field-of-View. Furthermore, the geometry of the environment, which is characterized by soft slopes and small objects, does not provide structures of great extent in the vertical direction as it is, instead, very common in datasets targeted at autonomous driving \cite{geiger2012we,maddern20171}. For these reasons, traditional LiDAR SLAM can not be employed successfully on this data. Compared to the stereo camera, however, where the depth uncertainty grows quadratically with the distance, the LiDAR sensor allows to measure accurately the shape of farther landscapes that might not be possible to reach, e.g., with a mobile system. It is, therefore, of great interest to develop SLAM pipelines where the LiDAR and visual sensing approaches are combined. 

Meant purely as an example, we developed a simple mapping pipeline, that assembles LiDAR scans synchronized to pose estimated from a Visual SLAM approach. For this, we used OPEN\_VINS. Fig.~\ref{fig:example_lidar_mapping} shows four examples of LiDAR maps from the \etnaone, \etnalandmarks, \etnastraight\ and \etnacistinout\ sequences, respectively. Observing these maps, it is evident that the predominant geometries are slopes, with the exception of sparse and relatively small rocks and stones. 
Other more evident structures, i.e., big rock formations,
are visible in Fig.\ref{fig:example_lidar_mapping}(b), which are re-observed twice. Fig~\ref{fig:example_lidar_mapping}(d), instead, shows the reconstruction of the rim of the Cisternazza crater, to an extent and detail that is not possible to achieve with a stereo camera alone. This dataset, therefore, intends to challenge visual-LiDAR SLAM algorithms to robustly perform pose estimation and mapping by developing novel approaches for place recognition that depart from traditional global LiDAR descriptors \cite{kim2018scan} or segmentation of structures with strong ground plane assumptions \cite{dube2018segmap}. 

\begin{figure}[t]
    \centering
    \subfloat{\includegraphics[width=.49\linewidth]{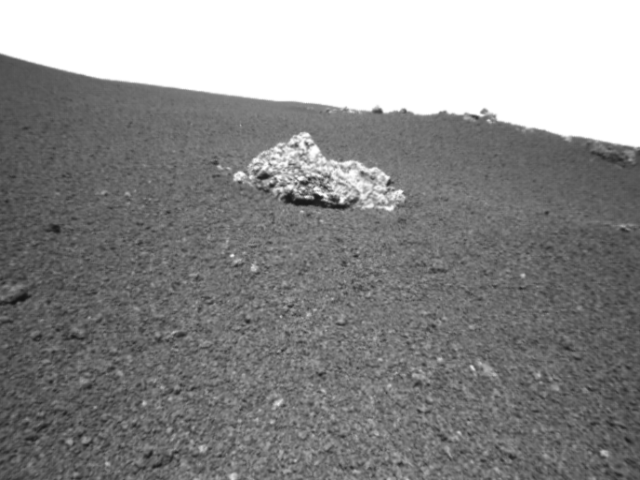}} \hfill
    \subfloat{\includegraphics[width=.49\linewidth]{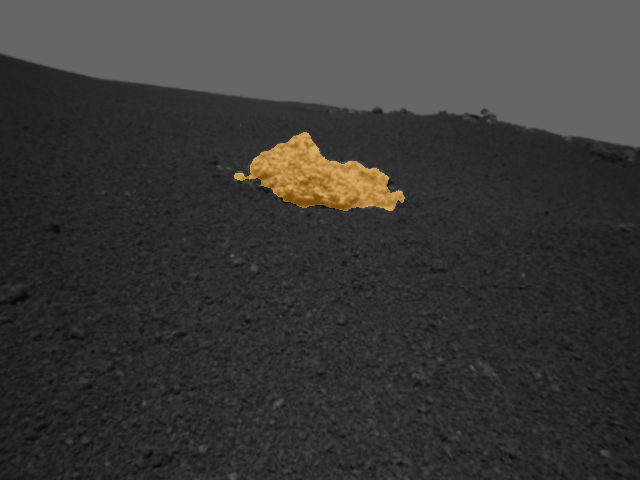}} \\[1mm] 
    \subfloat{\includegraphics[width=.49\linewidth, trim=0 0 0.05cm 0, clip]{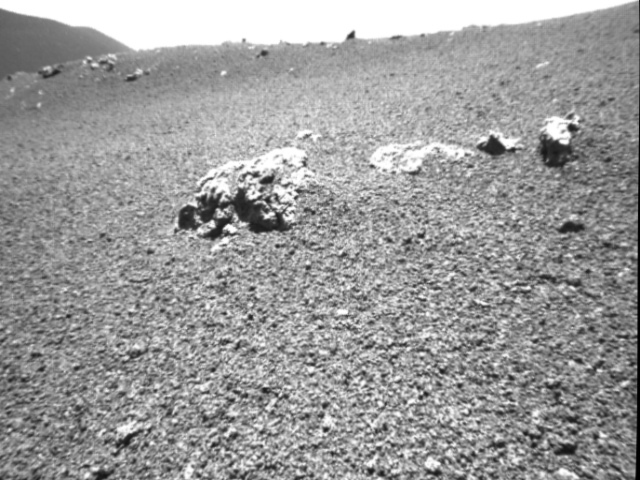}} \hfill
    \subfloat{\includegraphics[width=.49\linewidth, trim=0 0 0.05cm 0, clip]{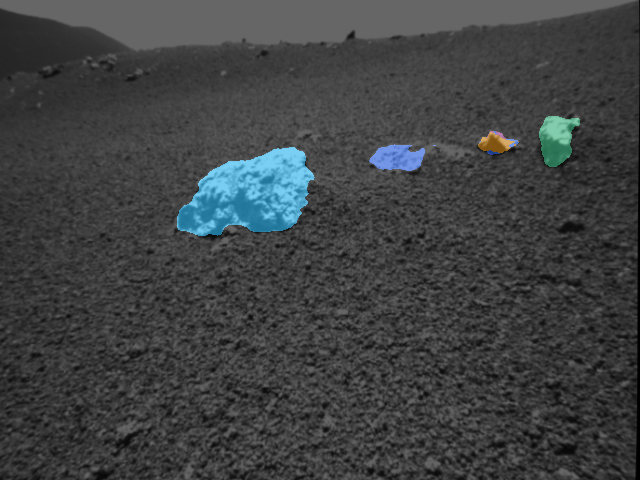}}
    \caption{Examples of instance segmentation of unknown objects, in this case small stones, using INSTR~\cite{durner2021unknown}. Shown are original left camera images (left column) and instance predictions as colored overlays (right column). 
    }
    \label{fig:instr_examples}
\end{figure}

In order to illustrate what may lead to such methods, we employed the Instance Stereo Transformer (INSTR)~\cite{durner2021unknown}, to detect instances of unknown objects in a scene (see Fig.~\ref{fig:instr_examples} for two examples). Note that the network has not been trained on this specific data. The segmentation masks can be used as an input for isolating objects in, otherwise complicated to process, LiDAR scans and accurately generate geometric models with the goal of realizing landmark-based \cite{dube2018segmap} or metric-semantic \cite{tian2022kimera} SLAM.

\section{CONCLUSIONS}
We presented a dataset, recorded in a planetary analogue environment, that comprises stereo, LiDAR and inertial measurements, and allows to assess localization and mapping frameworks under challenging visual and structural appearance. 
We, furthermore, evaluated a variety of visual-inertial SLAM algorithms on the sequences, highlighting strengths and limitations, and provided examples of potential use-cases for this data, that depart from traditional visual-LiDAR SLAM approaches.
This dataset, therefore, provides a mean to evaluate and overcome limitations of traditional SLAM algorithms, that are usually developed and tested on datasets which lack elements of visual aliasing and ambiguous geometries, typical of extremely unstructured environments. 


\section*{ACKNOWLEDGMENT}

This work was supported by the Helmholtz Association, project ARCHES (contract number ZT-0033). Special thanks to Florian Auinger for his work on the system, to Lukas Meyer for the support on the mountain and to Emanuel Staudinger for the help on the D-GNSS setup. 

\bibliographystyle{IEEEtran}
\bibliography{bibliography}

\begin{thebibliography}{10}
\providecommand{\url}[1]{#1}
\csname url@rmstyle\endcsname
\providecommand{\newblock}{\relax}
\providecommand{\bibinfo}[2]{#2}
\providecommand\BIBentrySTDinterwordspacing{\spaceskip=0pt\relax}
\providecommand\BIBentryALTinterwordstretchfactor{4}
\providecommand\BIBentryALTinterwordspacing{\spaceskip=\fontdimen2\font plus
\BIBentryALTinterwordstretchfactor\fontdimen3\font minus
  \fontdimen4\font\relax}
\providecommand\BIBforeignlanguage[2]{{%
\expandafter\ifx\csname l@#1\endcsname\relax
\typeout{** WARNING: IEEEtran.bst: No hyphenation pattern has been}%
\typeout{** loaded for the language `#1'. Using the pattern for}%
\typeout{** the default language instead.}%
\else
\language=\csname l@#1\endcsname
\fi
#2}}
\renewcommand\BIBentryALTinterwordstretchfactor{4}

\bibitem{bresson2017simultaneous}
Bresson, \emph{et~al.}, ``Simultaneous localization and mapping: A survey of
  current trends in autonomous driving,'' \emph{IEEE Transactions on
  Intelligent Vehicles}, vol.~2, no.~3, pp. 194--220, 2017.

\bibitem{burki2019vizard}
B{\"u}rki, \emph{et~al.}, ``{VIZARD}: Reliable visual localization for
  autonomous vehicles in urban outdoor environments,'' in \emph{IEEE
  Intelligent Vehicles Symposium (IV)}, 2019, pp. 1124--1130.

\bibitem{mascaro2021towards}
Mascaro, \emph{et~al.}, ``Towards automating construction tasks: Large-scale
  object mapping, segmentation, and manipulation,'' \emph{Journal of Field
  Robotics}, vol.~38, no.~5, pp. 684--699, 2021.

\bibitem{shu2021slam}
Shu, \emph{et~al.}, ``{SLAM} in the field: An evaluation of monocular mapping
  and localization on challenging dynamic agricultural environment,'' in
  \emph{IEEE/CVF Winter Conference on Applications of Computer Vision}, 2021,
  pp. 1761--1771.

\bibitem{oliveira2021advances}
Oliveira, \emph{et~al.}, ``Advances in agriculture robotics: A state-of-the-art
  review and challenges ahead,'' \emph{Robotics}, vol.~10, no.~2, p.~52, 2021.

\bibitem{sharafutdinov2021comparison}
Sharafutdinov, \emph{et~al.}, ``Comparison of modern open-source visual {SLAM}
  approaches,'' \emph{arXiv preprint arXiv:2108.01654}, 2021.

\bibitem{dellenbach2021s}
Dellenbach, \emph{et~al.}, ``What's in my {LiDAR} odometry toolbox?''
  \emph{arXiv preprint arXiv:2103.09708}, 2021.

\bibitem{shan2018lego}
Shan and Englot, ``{LeGO-LOAM}: Lightweight and ground-optimized lidar odometry
  and mapping on variable terrain,'' in \emph{IEEE/RSJ International Conference
  on Intelligent Robots and Systems (IROS)}, 2018, pp. 4758--4765.

\bibitem{giubilato2020GPGM}
Giubilato, \emph{et~al.}, ``{GPGM-SLAM:} a robust {SLAM} system for
  unstructured planetary environments with gaussian process gradient maps,''
  \emph{CoRR}, vol. abs/2109.06596, 2021.

\bibitem{gentil2020gaussian}
Le~Gentil, \emph{et~al.}, ``Gaussian process gradient maps for loop-closure
  detection in unstructured planetary environments,'' in \emph{IEEE/RSJ
  International Conference on Intelligent Robots and Systems (IROS)}, 2020, pp.
  1895--1902.

\bibitem{gelfand2003geometrically}
Gelfand, \emph{et~al.}, ``Geometrically stable sampling for the {ICP}
  algorithm,'' in \emph{International Conference on 3-D Digital Imaging and
  Modeling (3DIM)}, 2003, pp. 260--267.

\bibitem{geiger2013vision}
Geiger, \emph{et~al.}, ``Vision meets robotics: The {KITTI} dataset,''
  \emph{The International Journal of Robotics Research}, vol.~32, no.~11, pp.
  1231--1237, 2013.

\bibitem{maddern20171}
Maddern, \emph{et~al.}, ``1 year, 1000 km: The {Oxford} {RobotCar} dataset,''
  \emph{The International Journal of Robotics Research}, vol.~36, no.~1, pp.
  3--15, 2017.

\bibitem{choi2018kaist}
Choi, \emph{et~al.}, ``{KAIST} multi-spectral day/night data set for autonomous
  and assisted driving,'' \emph{IEEE Transactions on Intelligent Transportation
  Systems}, vol.~19, no.~3, pp. 934--948, 2018.

\bibitem{wenzel20204seasons}
Wenzel, \emph{et~al.}, ``{4Seasons}: A cross-season dataset for multi-weather
  {SLAM} in autonomous driving,'' in \emph{DAGM German Conference on Pattern
  Recognition}.\hskip 1em plus 0.5em minus 0.4em\relax Springer, 2020, pp.
  404--417.

\bibitem{sturm2012benchmark}
Sturm, \emph{et~al.}, ``A benchmark for the evaluation of {RGB-D} {SLAM}
  systems,'' in \emph{IEEE/RSJ International Conference on Intelligent Robots
  and Systems (IROS)}, 2012, pp. 573--580.

\bibitem{schubert2018tum}
Schubert, \emph{et~al.}, ``The {TUM} {VI} benchmark for evaluating
  visual-inertial odometry,'' in \emph{IEEE/RSJ International Conference on
  Intelligent Robots and Systems (IROS)}, 2018, pp. 1680--1687.

\bibitem{Schops_2019_CVPR}
Schops, \emph{et~al.}, ``{BAD SLAM}: Bundle adjusted direct {RGB-D} {SLAM},''
  in \emph{IEEE/CVF Conference on Computer Vision and Pattern Recognition
  (CVPR)}, 2019.

\bibitem{handa2014benchmark}
Handa, \emph{et~al.}, ``A benchmark for {RGB-D} visual odometry, {3D}
  reconstruction and {SLAM},'' in \emph{IEEE International Conference on
  Robotics and Automation (ICRA)}, 2014, pp. 1524--1531.

\bibitem{wang2020tartanair}
Wang, \emph{et~al.}, ``{TartanAir}: A dataset to push the limits of visual
  {SLAM},'' in \emph{IEEE/RSJ International Conference on Intelligent Robots
  and Systems (IROS)}, 2020, pp. 4909--4916.

\bibitem{furgale2012devon}
Furgale, \emph{et~al.}, ``The {Devon} {Island} rover navigation dataset,''
  \emph{The International Journal of Robotics Research}, vol.~31, no.~6, pp.
  707--713, 2012.

\bibitem{leung2017chilean}
Leung, \emph{et~al.}, ``Chilean underground mine dataset,'' \emph{The
  International Journal of Robotics Research}, vol.~36, no.~1, pp. 16--23,
  2017.

\bibitem{miller2018visual}
Miller, \emph{et~al.}, ``The visual--inertial canoe dataset,'' \emph{The
  International Journal of Robotics Research}, vol.~37, no.~1, pp. 13--20,
  2018.

\bibitem{hewitt2018katwijk}
Hewitt, \emph{et~al.}, ``The {Katwijk} beach planetary rover dataset,''
  \emph{The International Journal of Robotics Research}, vol.~37, no.~1, pp.
  3--12, 2018.

\bibitem{rogers2020test}
Rogers, \emph{et~al.}, ``Test your {SLAM}! the {SubT}-{Tunnel} dataset and
  metric for mapping,'' in \emph{IEEE International Conference on Robotics and
  Automation (ICRA)}, 2020, pp. 955--961.

\bibitem{bouman2020autonomous}
Bouman, \emph{et~al.}, ``Autonomous {Spot}: Long-range autonomous exploration
  of extreme environments with legged locomotion,'' in \emph{IEEE/RSJ
  International Conference on Intelligent Robots and Systems (IROS)}, 2020, pp.
  2518--2525.

\bibitem{rouvcek2019darpa}
Rou{\v{c}}ek, \emph{et~al.}, ``{DARPA} subterranean challenge: Multi-robotic
  exploration of underground environments,'' in \emph{International Conference
  on Modelling and Simulation for Autonomous Systems}, 2019, pp. 274--290.

\bibitem{meyer2021madmax}
Meyer, \emph{et~al.}, ``The {MADMAX} data set for visual-inertial rover
  navigation on {Mars},'' \emph{Journal of Field Robotics}, 2021.

\bibitem{vayugundla2018datasets}
Vayugundla, \emph{et~al.}, ``Datasets of long range navigation experiments in a
  {Moon} analogue environment on mount {Etna},'' in \emph{International
  Symposium on Robotics (ISR)}, 2018, pp. 1--7.

\bibitem{preston2012concepts}
Preston, \emph{et~al.}, ``{CAFE} -- concepts for activities in the field for
  exploration. {TN2}: The catalogue of planetary analogues,'' \emph{The Open
  University, Milton Keynes, UK}, 2012.

\bibitem{cramariuc2021semsegmap}
Cramariuc, \emph{et~al.}, ``{SemSegMap} -- {3D} segment-based semantic
  localization,'' in \emph{IEEE/RSJ International Conference on Intelligent
  Robots and Systems (IROS)}, 2021, pp. 1183--1190.

\bibitem{schuster2019towards}
Schuster, \emph{et~al.}, ``Towards autonomous planetary exploration,''
  \emph{Journal of Intelligent \& Robotic Systems}, vol.~93, no.~3, pp.
  461--494, 2019.

\bibitem{campos2021orb}
Campos, \emph{et~al.}, ``{ORB-SLAM3}: An accurate open-source library for
  visual, visual--inertial, and multimap {SLAM},'' \emph{IEEE Transactions on
  Robotics}, vol.~37, no.~6, pp. 1874--1890, 2021.

\bibitem{usenko19nfr}
Usenko, \emph{et~al.}, ``Visual-inertial mapping with non-linear factor
  recovery,'' \emph{IEEE Robotics and Automation Letters (RA-L) \&
  International Conference on Intelligent Robotics and Automation (ICRA)},
  vol.~5, no.~2, pp. 422--429, 2020.

\bibitem{Geneva2020ICRA}
Geneva, \emph{et~al.}, ``{OpenVINS}: A research platform for visual-inertial
  estimation,'' in \emph{IEEE International Conference on Robotics and
  Automation (ICRA)}, 2020.

\bibitem{qin2019a}
Qin, \emph{et~al.}, ``A general optimization-based framework for local odometry
  estimation with multiple sensors,'' 2019.

\bibitem{callab}
{S}trobl, \emph{et~al.} {D}{L}{R} {C}al{D}e and {D}{L}{R} {C}al{L}ab.
  \url{http://www.robotic.dlr.de/callab}. Institute of Robotics and
  Mechatronics, German Aerospace Center (DLR). Oberpfaffenhofen, Germany.

\bibitem{furgale2013unified}
Furgale, \emph{et~al.}, ``Unified temporal and spatial calibration for
  multi-sensor systems,'' in \emph{IEEE/RSJ International Conference on
  Intelligent Robots and Systems (IROS)}, 2013, pp. 1280--1286.

\bibitem{zhou2018automatic}
Zhou, \emph{et~al.}, ``Automatic extrinsic calibration of a camera and a {3D}
  {LiDAR} using line and plane correspondences,'' in \emph{IEEE/RSJ
  International Conference on Intelligent Robots and Systems (IROS)}, 2018, pp.
  5562--5569.

\bibitem{gong20133d}
Gong, \emph{et~al.}, ``{3D} {LIDAR}-camera extrinsic calibration using an
  arbitrary trihedron,'' \emph{Sensors}, vol.~13, no.~2, pp. 1902--1918, 2013.

\bibitem{toth2020automatic}
T{\'o}th, \emph{et~al.}, ``Automatic {LiDAR}-camera calibration of extrinsic
  parameters using a spherical target,'' in \emph{IEEE International Conference
  on Robotics and Automation (ICRA)}, 2020, pp. 8580--8586.

\bibitem{levinson2013automatic}
Levinson and Thrun, ``Automatic online calibration of cameras and lasers.'' in
  \emph{Robotics: Science and Systems}, 2013.

\bibitem{castorena2016autocalibration}
Castorena, \emph{et~al.}, ``Autocalibration of {LiDAR} and optical cameras via
  edge alignment,'' in \emph{IEEE International Conference on Acoustics, Speech
  and Signal Processing (ICASSP)}, 2016, pp. 2862--2866.

\bibitem{geiger2012we}
Geiger, \emph{et~al.}, ``Are we ready for autonomous driving? the {KITTI}
  vision benchmark suite,'' in \emph{IEEE Conference on Computer Vision and
  Pattern Recognition (CVPR)}, 2012, pp. 3354--3361.

\bibitem{kim2018scan}
Kim and Kim, ``Scan {C}ontext: Egocentric spatial descriptor for place
  recognition within {3D} point cloud map,'' in \emph{IEEE/RSJ International
  Conference on Intelligent Robots and Systems (IROS)}, 2018, pp. 4802--4809.

\bibitem{dube2018segmap}
Dub{\'e}, \emph{et~al.}, ``{SegMap}: {3D} segment mapping using data-driven
  descriptors,'' \emph{arXiv preprint arXiv:1804.09557}, 2018.

\bibitem{durner2021unknown}
Durner, \emph{et~al.}, ``Unknown object segmentation from stereo images,'' in
  \emph{IEEE/RSJ International Conference on Intelligent Robots and Systems
  (IROS)}, 2021, pp. 4823--4830.

\bibitem{tian2022kimera}
Tian, \emph{et~al.}, ``Kimera-multi: Robust, distributed, dense metric-semantic
  slam for multi-robot systems,'' \emph{IEEE Transactions on Robotics}, 2022.

\end{thebibliography}

\end{document}